\documentclass[
]{ceurart}

\usepackage{listings}
\begin{document}

\copyrightyear{2025}

\copyrightclause{Copyright © 2025 for this paper by its authors. Use permitted under Creative Commons License Attribution 4.0 International (CC BY 4.0).}

\conference{2nd International Workshop on Natural Scientific Language Processing and Research Knowledge Graphs (NSLP 2025), co-located with ESWC 2025, June 01–02, 2025, Portorož, Slovenia}

\title{ConExion: Concept Extraction with Large Language Models}


\author[1,2]{Ebrahim Norouzi}[%
orcid=0000-0002-2691-6995,
email=ebrahim.norouzi@fiz-karlsruhe.de,
]
\cormark[1]
\author[1]{Sven Hertling}[%
orcid=0000-0003-0333-5888,
email=sven.hertling@fiz-karlsruhe.de,
]

\author[1,2]{Harald Sack}[%
orcid=0000-0001-7069-9804,
email=harald.sack@fiz-karlsruhe.de,
]

\address[1]{FIZ Karlsruhe – Leibniz Institute for Information Infrastructure,\\
Hermann-von-Helmholtz-Platz 1, 76344 Eggenstein-Leopoldshafen, Germany}

\address[2]{Karlsruhe Institute of Technology (AIFB),\\
Kaiserstr. 89, 76133 Karlsruhe, Germany}

\cortext[1]{Corresponding author.}

\begin{abstract}
  In this paper, an approach for concept extraction from documents using pre-trained large language models (LLMs) is presented. Compared with conventional methods that extract keyphrases summarizing the important information discussed in a document, our approach tackles a more challenging task of extracting all present concepts related to the specific domain, not just the important ones. Through comprehensive evaluations of two widely used benchmark datasets, we demonstrate that our method improves the $F_1$ score compared to state-of-the-art techniques. Additionally, we explore the potential of using prompts within these models for unsupervised concept extraction. The extracted concepts are intended to support domain coverage evaluation of ontologies and facilitate ontology learning, highlighting the effectiveness of LLMs in concept extraction tasks. Our source code and datasets are publicly available at \url{https://github.com/ISE-FIZKarlsruhe/concept_extraction}.
\end{abstract}


\maketitle

\section{Introduction}
\label{sec:Introduction}

Concept/Keyword extraction is recognized as a fundamental task in Natural Language Processing (NLP), crucial for identifying and extracting noun phrases that summarize and represent the main topics discussed in a document \cite{survey2014}.
It has been extensively utilized in various applications, including information retrieval, text summarization, and text categorization \cite{kp20k}.

However, it is not only helpful in those fields of research but also in ontology evaluation methods.
Usually, competency questions (CQs) are created to specify the ontology requirements and are used later for the evaluation of the ontology~\cite{ont_eval_cq}.
In cases where the ontology is already developed and no CQs and domain experts are available, other approaches need to be applied.
\cite{raad2015survey} presents four approaches: (1) Gold Standard-based, (2) Corpus-based, (3) Task-based, and (4) Criteria based. 
The idea behind corpus-based methods is to compare the ontology with concepts extracted from a text corpus that significantly covers the given domain.
Thus, a good approach is needed to extract those concepts from a given text.

The task of concept extraction can be formally defined as follows: Given a document $D$ represented as a sequence of words $D = [w_1, w_2, w_3,..., w_n]$,  the goal is to extract a set $C={(c_1, s_1), (c_2, s_2), ..., (c_m, s_m)}$ where $c_i$ consists of one or multiple words that best represent the topics of the document $D$. The additional score $s_i\in[0,1]$ provides a confidence value of the approach to be able to specify the importance of each extracted concept further. The score can also be used to rank the phrases and only extract the top $p$ concepts.

Concept extraction is typically categorized into two types: (i) extracting present keyphrases (extractive), which are directly found within the input document such that $k_i\in D$, and (ii) extracting absent keyphrases (abstractive), which are generated even though they do not appear explicitly in the document \cite{survey2023}.

In this paper, we concentrate on extracting present concepts from documents using large language models (LLMs) because this approach best fits the task of ontology evaluation. Ontology evaluation \cite{gomez2004ontology} methods are used to judge an ontology’s content with respect to a reference framework throughout its development lifecycle. Since ontology evaluation includes verification and validation, it’s important to trace each extracted concept back to where it appears in the original reference framework. Extractive methods naturally fulfill this requirement by ensuring that all identified concepts are grounded in the original document, enabling a more accurate assessment of an ontology’s coverage and relevance.
The approaches are evaluated on two common datasets for concept extraction, which are about scientific publications.
We show that with a simple yet powerful and reproducible setup, we can surpass the state-of-the-art approaches.

Our contributions include:
\begin{itemize}
    \item Utilizing large language models for effective concept extraction from scientific documents.
    \item Providing comprehensive evaluations on benchmark datasets.
    \item Demonstrating improved performance over state-of-the-art techniques in extracting present concepts.
\end{itemize}

The paper is structured as follows: Section \ref{sec:Introduction} introduces the problem and outlines the objectives of this study. Section \ref{sec:related_work} reviews the existing literature on concept extraction and large language models, highlighting the gap this study aims to fill. Section \ref{sec:methodology} details the methodology, the experimental setup, and the various prompts tested. Section \ref{sec:Evaluation} presents the evaluation metrics, the datasets used, and the results of the experiments, with a detailed analysis of the performance of different models and prompts. Section \ref{sec:Conclusion} concludes the paper, summarizing the key findings and outlining directions for future work.

\section{Related Work}
\label{sec:related_work}


Several comprehensive surveys provide a thorough overview of concept extraction techniques \cite{survey2014,survey2023,song-etal-2023-survey}. This paper focuses specifically on unsupervised methods, which offer several advantages such as domain independence and the lack of a requirement for training data. These methods are currently recognized as state-of-the-art in terms of performance.

Recent advancements in unsupervised concept extraction have led to the development of several innovative approaches that do not rely on annotated data. Statistical methods, such as TF-IDF \cite{tfidf} and YAKE \cite{yake}, identify significant words by calculating statistical features like word frequencies and co-occurrences, thereby selecting candidates for keyphrases.

\paragraph{Graph-based methods} have also been widely explored. TextRank \cite{TextRank} represents text as a graph, where words serve as nodes and their co-occurrences act as edges. These methods employ node ranking algorithms, such as PageRank \cite{PageRank}, to rank words and extract the top-k words as keyphrase candidates. Bougouin et al. \cite{TopicRank} introduced TopicRank, which clusters candidate phrases into topics in the initial phase and subsequently ranks these topics based on their significance within the document. Positionrank \cite{PositionRank} further enhances this approach by incorporating the positional information of words in the text to improve ranking accuracy.

\paragraph{Embedding-based models} have emerged as another effective approach for key\-phrase extraction. EmbedRank \cite{EmbedRank}, for instance, utilizes part-of-speech tagging to identify potential keyphrases from a document. These keyphrases are then represented as low-dimensional vectors using a pretrained embedding model and ranked according to their Cosine similarity to the document’s overall embedding vector. This technique leverages the semantic representations provided by embedding models to improve keyphrase extraction. MultPAX \cite{MultPAX} employs semantic similarity between candidate keyphrases and an input document using the pretrained embedding of BERT model. AutoKeyGen \cite{AutoKeyGen} constructs a phrase bank by combining keyphrases from all documents into a corpus, considering both lexical and semantic similarities for selecting top candidate keyphrases for each input document. PromptRank \cite{PromptRank} is an unsupervised approach based on a pre-trained language model (PLM) with an encoder-decoder architecture. PromptRank feeds the document into the encoder and calculates the probability of generating the candidate with a designed prompt by the decoder to rank the candidates. A limitation of PromptRank is that its performance heavily depends on the design of prompts and may not always guarantee optimal results.

Recent efforts on pre-trained large language models, such as ChatGPT and ChatGLM, have demonstrated promising performance using zero-shot prompts, inspiring the exploration of prompt-based methods for keyphrase extraction \cite{song2024large}. However, these efforts are limited to ChatGPT and ChatGLM in zero-shot prompts. While these models perform well in various NLP tasks without parameter tuning, experimental results indicate that ChatGPT has significant room for improvement in keyphrase extraction compared to state-of-the-art models. This gap is attributed to limited resource settings and basic experimental configurations. More sophisticated methods, including complex prompt designs and contextual sample construction, are necessary to optimize its performance. A comprehensive study reports that models like GPT-3, InstructGPT, and ChatGPT show modest improvements in keyphrase predictions \cite{survey2023}. Furthermore, existing research has primarily focused on two zero-shot prompts (asking for extraction of keywords and keyphrases from documents) and limited few-shot configurations (1-shot and 5-shot asking to extract keywords), indicating a need for a more comprehensive study of open-source LLM models with various prompts.

\section{LLMs for Concept Extraction}
\label{sec:methodology}

\begin{figure}[t]
    \centering
    \includegraphics[width=\textwidth]{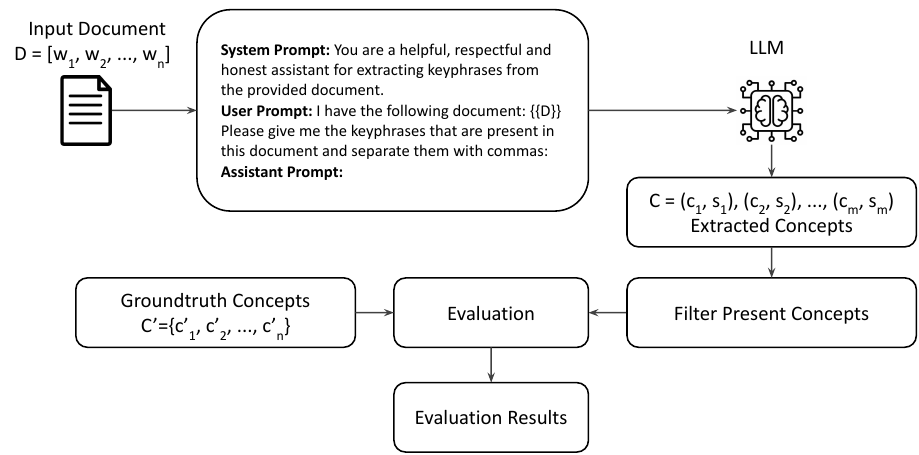}
    \caption{Overall approach of ConExion.}
    \label{fig:llm-approach}
\end{figure}

The concept extraction approach uses various Large Language Models (LLMs) to identify and extract keywords from input documents. Figure \ref{fig:llm-approach} shows the workflow. Initially, an input document is provided containing the text from which concepts need to be extracted. This document and a specific prompt are used for the extraction process. The prompts include a system prompt, a user prompt, and an assistant prompt. The system prompt sets the context for the extraction task. The user prompt directly asks the model to perform the extraction, and the assistant prompt is where the LLM outputs the extracted keywords.
Before the evaluation, the extracted concepts are further filtered because only concepts present in the input document should be returned. Thus, it is easily possible to check if each extracted concept is actually contained in the document and filter those that are not.

The LLM will generate natural language text as an answer. Even though it is asked to only return the concepts separated with commas, various conversational statements can appear at the beginning of the generated text, like "Sure, I'd be happy to help! Based on the information provided in the document ...". Another issue is that the output format does not always need to fit the requested format which should separate the concepts with commas. 

To overcome all these problems, the extraction of the concepts, given the produced text, works as follows: 
The text is split by tokens ",", ";", "*", and "\textbackslash n"(newline). The reason for the first two is that the LLM might separate the concepts not only by a comma but also by a semicolon. The star is used as a separator because some models might also return a markdown formatted unordered list. Finally, the newline is also included because conversational sentences or phrases are suffixed with a newline to nicely format the output (most probably due to fine-tuning to a chat-style). The keywords that only appear in the document are selected. Thus also those conversational sentences will be removed.

The task of concept extraction involves not only to extract the concepts themselves but also an attached confidence score to finally rank the extracted concepts.
An LLM usually does not provide a confidence score in the generated output even though it is asked for it because, on the one hand, it is difficult to generate numbers that reflect the confidence of the tokens previously generated, and on the other hand, the training data usually prevent the model from generating such numbers (the model often outputs sentences such as "As an LLM, I'm not able to provide confidence scores").
Therefore we implemented another approach to extract the confidence scores of the model.
When generating a new token, the model computes a probability distribution over their entire vocabulary, where each token is assigned a likelihood based on the model’s internal scoring function. To extract a confidence score for a generated concept, we retrieve the probability assigned to each token in the phrase during generation and compute the geometric mean of these probabilities, providing a single aggregated confidence value.
There are several generation strategies for how the next token is then generated. Greedy search, for example,
always selects the token with the highest probability such that the results are always the same (given the same input tokens).
Other approaches like multinominal sampling or contrastive search \cite{su2022contrastive} are non-deterministic because they sample the next token based on the computed distribution and thus could generate different results.
For reproducibility reasons, we selected the greedy search and finally computed a confidence score for each extracted concept by multiplying the probabilities of all tokens that form this concept. 


\begin{table}[tp]
    \centering
    \caption{Prompt templates for extracting concepts.}
    \label{tab:prompt_templates}
    \resizebox{\textwidth}{!}{
    \begin{tabular}{m{0.2\textwidth}|m{0.9\textwidth}}
        \hline
        \textbf{Name} & \textbf{Prompt} \\
        \hline
        \multicolumn{2}{c}{Zero-Shot (ZS)}\\ \hline
        ZS Keywords & 
        \textbf{User:} Please give me the keywords that are present in this document and separate them with commas: 
        \textbf{Assistant: } \\
        \hline
        ZS Keyphrases & 
        \textbf{User:} Please give me the keyphrases that are present in this document and separate them with commas: 
        \textbf{Assistant: } \\
        \hline
        ZS Concepts & 
        \textbf{User:} Please give me the concepts that are present in this document and separate them with commas: 
        \textbf{Assistant: } \\
        \hline
        ZS Entities & 
        \textbf{User:} Please give me the entities that are present in this document and separate them with commas: 
        \textbf{Assistant: } \\
        \hline
        ZS Topics & 
        \textbf{User:} Please give me the topics that are present in this document and separate them with commas: 
        \textbf{Assistant: } \\
        \hline
        \multicolumn{2}{c}{Zero-Shot with more domain information}\\ \hline
        ZS + Domain & 
        \textbf{User:} Please give me the keyphrases related to the domains of Computer Science, Control, and Information Technology that are present in this document and separate them with commas: 
        \textbf{Assistant: } \\
        \hline
        \multicolumn{2}{c}{Zero-Shot with situational context}\\ \hline
        ZS + Extracting Context & 
        \textbf{System:} You are a helpful, respectful and honest assistant for extracting keyphrases from the provided document. 
        \textbf{User:} I have the following document: [DOCUMENT] Please give me the keyphrases that are present in this document and separate them with commas: 
        \textbf{Assistant: } \\
        \hline
        ZS + Expert Context & 
        \textbf{System:} You are an ontology expert in extracting keyphrases from the document. 
        \textbf{User:} I have the following document: [DOCUMENT] Please give me the keyphrases that are present in this document and separate them with commas: 
        \textbf{Assistant: } \\
        \hline
        \multicolumn{2}{c}{Zero-Shot with task description}\\ \hline
        ZS + Task Context & 
        \textbf{System:} You are an expert in extracting keyphrases from documents. Keyphrases are important multi- or single noun phrases that cover main topics of the document. 
        \textbf{User:} I have the following document: [DOCUMENT] Please give me the Keyphrases that are present in this document and separate them with commas: 
        \textbf{Assistant: } \\
        \hline
        \multicolumn{2}{c}{Few-Shot (FS)}\\
        \hline
        \multirow{3}{*}{FS n-Fixed} & \multirow{3}{11cm}{For each training document $D_i$ and groundtruth keyphrases $K_i$ ($i = 1$ to $n$):
            \begin{itemize}
                \item[] \textbf{User:} I have the following document: $D_i$
                \item[] Please give me the keyphrases that are present in this document and separate them with commas:
                \item[] \textbf{Assistant:} $K_i$
            \end{itemize}
            \textbf{User:} Please give me the keyphrases that are present in this document and separate them with commas: \\
            \textbf{Assistant:}
        } \\
        & \\
        & \\
        \cline{1-1}
        \multirow{3}{*}{FS n-Random} & \\
        & \\
        & \\
        \cline{1-1}
        \multirow{3}{*}{FS n-Closest} & \\
        & \\
        & \\
        \hline

    \end{tabular}
    }
\end{table}

For the prompt design, we chose to make a systematic analysis.  
To design the prompt configurations, various search terms, and prompt setups were explored for effective concept extraction.
We first started with a simple prompt (see \texttt{ZS Keywords} in Table \ref{tab:prompt_templates}) but modified the search term to identify the most suitable terminology. The search term is the word that is used to describe the task, e.g., "give me the keywords/concepts/entities in this document," where keywords, concepts, and entities are the search terms. Overall, five of those search terms are tried out, and the best-performing word is selected.

All other prompts are based on the previously selected prompt but extended in various ways.
This includes prompts with more precise domain information, situational contexts, and detailed task descriptions. For example, the \texttt{ZS + Domain prompt} includes more precise domain information, guiding the model to extract keyphrases related to specific fields such as Computer Science, Control, and Information Technology (which are the topics of the selected datasets). The \texttt{ZS + Extracting Context} prompt asks the model to act as a helpful assistant specifically for extracting keyphrases. The \texttt{ZS + Expert Context} prompt requires the model to take on the role of an ontology expert to extract keyphrases. Lastly, the \texttt{ZS + Task Context} prompt provides a detailed description of the extraction task, outlining the description of keyphrases, and asking the model to identify these keyphrases within the document.
All previously mentioned prompts do not use any training example and are thus also called zero-shot (ZS) prompts.

Few-shot (FS) prompts, on the other hand, utilize a specific number of examples from the training data to guide the extraction process. This study uses prompts with one, three, and five examples. These examples can be fixed, randomly selected, or chosen based on the closest embeddings to the training data. The \texttt{n-Fixed} prompts use n training examples sampled from the training data and those are fixed during all test examples whereas \texttt{n-Random} prompts use n randomly selected examples from the training data for each document in the test set. The \texttt{n-Closest} prompts select the top n examples in the training set based on the topical similarity of the document that needs to be predicted. Computing the similarity measure is done by embedding the training corpus using Sentence-BERT \cite{reimers-2019-sentence-bert} and retrieving the top-n closest documents given the embedded prediction document using cosine similarity. The model 'all-mpnet-base-v2' is used because of its superior quality across 14 diverse tasks from different domains\footnote{\url{https://www.sbert.net/docs/pretrained_models.html}}. These categorically different prompts allow for a flexible and comprehensive approach to concept extraction, accommodating varying levels of context and specificity required for different tasks.

Since each LLM model is trained on datasets specifically formatted as a chat history with user and assistant roles, each model needs a different chat template e.g. a prompt for Llama2 model needs to start with a special token "<s>" and each user prompt is encapsulated in instruction tokens "[INST]" and "[/INST]".
Thus, only the system, user, and assistant phrases are defined, and the final prompt is generated by applying the correct chat template for the model currently being executed. This is done by executing the \texttt{apply\_chat\_template} function of the Huggingface tokenizer\footnote{\url{https://huggingface.co/docs/transformers/chat_templating}}.

\section{Evaluation}
\label{sec:Evaluation}

In this section, the ConExion approach is evaluated on two common datasets.
All experiments are executed on a server running RedHat with 256 GB of RAM, 2x64 CPU
cores (2.6 GHz), and 2 Nvidia A100 (40GB) graphics cards (depending on the model size, only 1 GPU is used).

\subsection{Dataset}
\label{sec:evaluation_datasets}

\begin{table}[tp]
    \centering
    \caption{Statistics of publicly available datasets used for evaluation. The columns are defined as follows: $N_{doc}$: Number of documents, $Avg_{doc}$: Average document length (words), $Max_{doc}$: Maximum document length (words), $Max_{con}$: Maximum number of concepts per documents, $Min_{con}$: Minimum number of concepts per documents, and $Avg_{con}$: Average number of concepts per documents.}
    \resizebox{\textwidth}{!}{
    \begin{tabular}{>{\centering\arraybackslash}p{1.9cm}|>{\centering\arraybackslash}p{.7cm}>{\centering\arraybackslash}p{1cm}>{\centering\arraybackslash}p{1cm}>{\centering\arraybackslash}p{1cm}>{\centering\arraybackslash}p{2.4cm}>{\centering\arraybackslash}p{0.7cm}>{\centering\arraybackslash}p{0.7cm}>{\centering\arraybackslash}p{0.7cm}>{\centering\arraybackslash}p{0.7cm}>{\centering\arraybackslash}p{0.7cm}}
        \hline
        \textbf{Dataset} & \textbf{Set} & \boldmath{$N_{doc}$} & \boldmath{$Avg_{doc}$} & \boldmath{$Max_{doc}$} & \boldmath{$Max_{con}$} / \boldmath{$Min_{con}$} / \boldmath{$Avg_{con}$} & \multicolumn{5}{r}{\textbf{Concepts Distribution}} \\
        \cline{7-11}
        & & & & & & \textbf{1} & \textbf{2} & \textbf{3} & \textbf{4} & \textbf{$\geq$5} \\
        \hline
        \multirow{2}{*}{Inspec} & Train & 976 & 143.35 & 557 & 24 / 1 / 5.38 & 14.3 & 55.3 & 23.9 & 5.3 & 1.1 \\
        & Test & 486 & 135.73 & 384 & 25 / 1 / 5.55 & 14.3 & 58.3 & 22.5 & 3.9 & 1.0 \\
        \hline
        \multirow{2}{*}{SemEval2017} & Train & 350 & 160.51 & 355 & 28 / 2 / 9.44 & 24.8 & 34.9 & 18.9 & 8.8 & 12.6 \\
        & Test & 100 & 190.40 & 297 & 23 / 1 / 8.72 & 29.1 & 44.5 & 13.4 & 6.4 & 6.5 \\
        \hline
    \end{tabular}
    }
    \label{tab:dataset_statistics}
\end{table}
This study utilizes two publicly available datasets: Inspec \cite{inspec}, and SemEval2017 \cite{semeval2017}. For each dataset, documents without ground truth keyphrases were filtered out, and the statistics presented in Table \ref{tab:dataset_statistics} are based on the filtered documents used in our approach. 
The Inspec\footnote{\url{https://huggingface.co/datasets/midas/inspec}} dataset comprises abstracts from 2,000 English scientific papers in the domains of Computers, Control, and Information Technology, published between 1998 and 2002. The SemEval-2017\footnote{\url{https://huggingface.co/datasets/midas/semeval2017}} dataset contains abstracts from 500 English scientific papers from ScienceDirect open access publications. The papers topics span computer/material science and physics. Concepts were annotated by student volunteers and subsequently validated by expert annotators.
The training and testing split was included to ensure a fair comparison with other traditional models. Additionally, examples from the training data were utilized in the prompts for few-shot learning purposes, where the model generates outputs based on contextual examples provided during prompting.

\subsection{Evaluation Metrics}
\label{sec:evaluation_metrics}
This section outlines the metrics used for evaluating concept extraction systems and presents the results on commonly-used datasets.
The performance of our concept extraction approach is assessed using standard metrics: Precision, Recall, and $F_1$-score \cite{AutoKeyGen}. Precision is calculated as the proportion of correctly predicted concepts out of all predicted concepts. Recall measures the proportion of correctly predicted concepts relative to the total number of concepts in the ground truth. $F_1$-score is the harmonic mean of Precision and Recall.

Many other works only report precision, recall, $F_1$-score at $k$ whereas $k$ is usually set to 5, 10, and 15 \cite{kong2023promptrankunsupervisedkeyphraseextraction,kp20k,song-etal-2023-survey,survey2023,MultPAX}. Given that the datasets usually contain less than nine keywords on average, those measures do not represent meaningful information (especially $F_1@15$ does not make any sense).
Nevertheless, we provide the values for $F_1@5$ and $F_1@10$ only for comparison possibilities to related work and to estimate the quality of the attached confidence scores. Again, due to compatibility with other works, the extracted concepts are further stemmed (but only for the scores at $k$) using the Porter Stemmer from NLTK library\footnote{\url{https://github.com/nltk/nltk/blob/develop/nltk/stem/porter.py}}.

Precision, recall, $F_1$-score at $k$ are computed as follows given a list of predicted concepts \( C = (c_1, c_2, \ldots, c_m) \) and target concepts \( C' = (c'_1, c'_2, \ldots, c'_n) \):

The Precision at top-$k$ (\( P@k \)) is defined as:
\[ P@k = \frac{|C:k \cap C'|}{|C:k|} \]

Recall (\( R@k \)) measures the proportion of correctly matched concepts among all ground truth concepts:
\[ R@k = \frac{|C:k \cap C'|}{|C':k|} \]

The \( F_1 \)-score at top-$k$ (\( F_1@k \)) is the harmonic mean of \( P@k \) and \( R@k \):
\[ F_1@k = \frac{2 \cdot P@k \cdot R@k}{P@k + R@k} \]

whereas (\( C:k \)) represents the top $k$ concepts according to the confidence score.

Especially for the use case of ontology evaluation, not the top-$k$ concepts are of interest but all concepts that are relevant to the document. Thus, we cannot rely on methods that only do the ranking of potentially many concepts but leave the decision of a good threshold to the user. As a result, only precision, recall, and $F_1$ are important and meaningful measures in this context.

\subsection{Selected Large Language Models}

In these experiments, we evaluated a diverse set of large language models to ensure coverage across both open and closed models, varying in scale and architectural design. Specifically, we selected: Llama-2-7b-chat-hf\footnote{\url{https://huggingface.co/meta-llama/Llama-2-7b-chat-hf}}, Llama-2-13b-chat-hf\footnote{\url{https://huggingface.co/meta-llama/Llama-2-13b-chat-hf}}, Llama-2-70b-chat-hf\footnote{\url{https://huggingface.co/meta-llama/Llama-2-70b-chat-hf}}, Llama-3-8B-Instruct\footnote{\url{https://huggingface.co/meta-llama/Meta-Llama-3-8B-Instruct}}, Llama-3-70B-Instruct\footnote{\url{https://huggingface.co/meta-llama/Meta-Llama-3-70B-Instruct}}, Mistral-7B-Instruct-v0.3\footnote{\url{https://huggingface.co/mistralai/Mistral-7B-Instruct-v0.3}}, Mixtral-8x7B-Instruct-v0.1\footnote{\url{https://huggingface.co/mistralai/Mixtral-8x7B-Instruct-v0.1}}, as well as GPT-3.5 Turbo\footnote{\url{https://platform.openai.com/docs/models/gpt-3-5-turbo}}.

\subsection{Evaluation Results}
\label{sec:evaluation_results}

\begin{table}[tp]
\centering
\caption{Precision, Recall, $F_1$, $F_1$@5, $F_1$@10, and Average Number of Extracted Concepts ($N_{EX}$) for Inspec dataset. Best results are highlighted in bold.}
\label{tab:zs_evaluations}
\begin{tabular}{l|lrrrrrr}
\hline
\textbf{Prompt} & \textbf{Model} & \textbf{P} & \textbf{R} & \textbf{\boldmath{$F_1$}} & \textbf{\boldmath{$F_1$@5}} & \textbf{\boldmath{$F_1$@10}} & \textbf{\boldmath{$N_{EX}$}} \\
\hline
\multirow{8}{*}{ZS Keywords}
&Llama2 7B & 0.162 & 0.358 & 0.208 & 0.164 & 0.220 & 13.889 \\
&Llama2 13B & 0.273 & 0.475 & 0.324 & 0.274 & 0.331 & 9.340 \\
&Llama2 70B & 0.178 & 0.487 & 0.239 & 0.235 & 0.297 & 17.675 \\ \cline{2-8}
&Llama3 8B & 0.208 & 0.473 & 0.270 & 0.199 & 0.257 & 19.056 \\
&Llama3 70B & 0.295 & 0.561 & 0.362 & 0.298 & 0.366 & 10.282 \\ \cline{2-8}
&Mistral 7B & 0.188 & 0.463 & 0.246 & 0.218 & 0.261 & 10.811 \\
&Mixtral 8x7B & 0.255 & 0.616 & 0.333 & 0.267 & 0.331 & 13.551 \\ \cline{2-8}
&GPT 3.5-turbo & 0.272 & \textbf{0.71} & 0.369 & 0.291 & 0.354 & 15.146 \\
\hline

\multirow{8}{*}{ZS Keyphrases}
&Llama2 7b & 0.207 & 0.419 & 0.256 & 0.230 & 0.278 & 9.154 \\
&Llama2 13b & 0.193 & 0.297 & 0.214 & 0.200 & 0.220 & 4.821 \\
&Llama2 70b & 0.235 & 0.492 & 0.293 & 0.283 & 0.355 & 11.619 \\ \cline{2-8}
&Llama3 8B & 0.297 & 0.597 & 0.373 & 0.313 & 0.379 & 10.922 \\
&Llama3 70B & \textbf{0.354} & 0.591 & \textbf{0.414} & \textbf{0.362} & \textbf{0.422} & 8.582 \\ \cline{2-8}
&Mistral 7B & 0.216 & 0.464 & 0.272 & 0.238 & 0.286 & 12.883 \\
&Mixtral 8x7B & 0.276 & 0.568 & 0.342 & 0.296 & 0.349 & 11.516 \\ \cline{2-8}
&GPT 3.5-turbo & 0.289 & 0.686 & 0.383 & 0.312 & 0.378 & 13.521 \\
\hline

\multirow{8}{*}{ZS Concepts}
&Llama2 7b & 0.144 & 0.261 & 0.173 & 0.166 & 0.203 & 5.977 \\
&Llama2-13b & 0.061 & 0.077 & 0.062 & 0.092 & 0.100 & 2.093 \\
&Llama2 70b & 0.226 & 0.437 & 0.274 & 0.255 & 0.318 & 10.331 \\ \cline{2-8}
&Llama3 8B & 0.292 & 0.609 & 0.371 & 0.309 & 0.369 & 11.918 \\
&Llama3 70B & 0.292 & 0.504 & 0.349 & 0.306 & 0.361 & 8.362 \\ \cline{2-8}
&Mistral 7B & 0.037 & 0.033 & 0.026 & 0.024 & 0.029 & 0.971 \\
&Mixtral 8x7B & 0.274 & 0.563 & 0.344 & 0.282 & 0.346 & 11.665 \\ \cline{2-8}
&GPT 3.5-turbo & 0.301 & 0.630 & 0.384 & 0.321 & 0.386 & 11.498 \\
\hline

\multirow{8}{*}{ZS Entities}
&Llama2 7b & 0.179 & 0.371 & 0.224 & 0.201 & 0.250 & 11.432 \\
&Llama2 13b & 0.102 & 0.149 & 0.111 & 0.125 & 0.144 & 3.889 \\
&Llama2 70b & 0.173 & 0.407 & 0.226 & 0.201 & 0.273 & 12.553 \\ \cline{2-8}
&Llama3 8B & 0.195 & 0.414 & 0.250 & 0.177 & 0.244 & 11.944 \\
&Llama3 70B & 0.192 & 0.348 & 0.233 & 0.192 & 0.235 & 9.068 \\ \cline{2-8}
&Mistral 7B & 0.144 & 0.252 & 0.165 & 0.140 & 0.167 & 5.889 \\
&Mixtral 8x7B & 0.231 & 0.519 & 0.298 & 0.236 & 0.296 & 12.469 \\ \cline{2-8}
&GPT 3.5-turbo & 0.263 & 0.662 & 0.356 & 0.282 & 0.351 & 14.195 \\ 
\hline

\multirow{8}{*}{ZS Topics}
&Llama2 7b & 0.092 & 0.144 & 0.100 & 0.114 & 0.131 & 3.967 \\
&Llama2 13b & 0.106 & 0.109 & 0.095 & 0.114 & 0.126 & 2.469 \\
&Llama2 70b & 0.193 & 0.336 & 0.221 & 0.230 & 0.276 & 8.401 \\ \cline{2-8}
&Llama3 8B & 0.323 & 0.601 & 0.394 & 0.328 & 0.399 & 10.247 \\
&Llama3 70B & 0.187 & 0.239 & 0.195 & 0.199 & 0.225 & 4.156 \\ \cline{2-8}
&Mistral 7B & 0.024 & 0.010 & 0.012 & 0.011 & 0.012 & 0.566 \\
&Mixtral 8x7B & 0.266 & 0.496 & 0.322 & 0.270 & 0.332 & 9.947 \\ \cline{2-8}
&GPT 3.5-turbo & 0.214 & 0.416 & 0.267 & 0.236 & 0.282 & 8.377 \\
\hline
\end{tabular}
\end{table}

\begin{table}[tp]
\centering
\caption{Precision, Recall, $F_1$, $F_1$@5, $F_1$@10, and Average Number of Extracted Keyphrases ($N_{EX}$) for Inspec dataset and Llama3 70B model. Best results are highlighted in bold.}
\label{tab:zs_fs_evaluations}
\begin{tabular}{l|rrrrrr}
\hline
\textbf{Prompt} & \textbf{P} & \textbf{R} & \textbf{\boldmath{$F_1$}} & \textbf{\boldmath{$F_1$@5}} & \textbf{\boldmath{$F_1$@10}} & \textbf{\boldmath{$N_{EX}$}} \\

\hline
\multicolumn{7}{c}{Zero-Shot}\\ \hline
\multirow{1}{*}{ZS + Domain}
& 0.299 & 0.350 & 0.291 & 0.286 & 0.308 & 5.274 \\ \hline

\multirow{1}{*}{ZS + Extracting Context}
& 0.367 & 0.600 & 0.428 & 0.372 & 0.432 & 8.558 \\ \hline

\multirow{1}{*}{ZS + Expert Context}
& 0.360 & \textbf{0.618} & 0.429 & 0.373 & 0.434 & 8.881 \\ \hline

\multirow{1}{*}{ZS + Task Context}
& 0.367 & 0.606 & 0.431 & 0.377 & 0.439 & 8.560 \\ \hline

\multicolumn{7}{c}{Few-Shot}\\ \hline
\multirow{1}{*}{FS 1-Fixed}
& 0.407 & 0.564 & 0.442 & 0.400 & 0.446 & 7.091 \\
\multirow{1}{*}{FS 3-Fixed}
& 0.400 & 0.568 & 0.440 & 0.394 & 0.446 & 7.189 \\
\multirow{1}{*}{FS 5-Fixed}
& 0.415 & 0.481 & 0.416 & 0.396 & 0.420 & 5.796 \\ \hline

\multirow{1}{*}{FS 1-Random}
& 0.413 & 0.575 & \textbf{0.451} & \textbf{0.406} & \textbf{0.453} & 7.327 \\
\multirow{1}{*}{FS 3-Random}
& 0.411 & 0.506 & 0.422 & 0.398 & 0.428 & 6.150 \\
\multirow{1}{*}{FS 5-Random}
& 0.416 & 0.478 & 0.415 & 0.395 & 0.422 & 5.790 \\ \hline

\multirow{1}{*}{FS 1-Closest}
& 0.410 & 0.569 & 0.445 & 0.397 & 0.446 & 7.340 \\
\multirow{1}{*}{FS 3-Closest}
& 0.411 & 0.499 & 0.421 & 0.392 & 0.429 & 6.185 \\
\multirow{1}{*}{FS 5-Closest}
& \textbf{0.426} & 0.497 & 0.429 & 0.402 & 0.435 & 5.879 \\
\hline
\end{tabular}
\end{table}

The evaluation of our approach was conducted using various types of prompts and different large language models (LLMs) to extract concepts from the Inspec dataset. The performance metrics considered are Precision, Recall, $F_1$-score, $F_1@5$, $F_1@10$. The average number of extracted concepts ($N_{EX}$) is included as well. This metric provides insight into how well the model aligns with the expected number of concepts typically found in the dataset. By comparing the average number of extracted concepts to the average number of present concepts in the dataset, we can better understand the model's extraction behavior and its tendency to over-extract or under-extract concepts. In the Inspec test dataset, the average number of concepts is 5.55. Therefore, reporting $N_{EX}$ helps in evaluating whether the model's output is consistent with this benchmark, ensuring a meaningful comparison of extraction performance across different models and prompts.




Table \ref{tab:zs_evaluations} presents the results for different zero-shot (ZS) prompts on the Inspec dataset. In these experiments, we tested several variations of the search terms (e.g., keywords, keyphrases, concepts, entities, topics) to see how different prompt wordings affected performance. These were selected based on their frequent appearance in prior literature on keyphrase and concept extraction. The results indicate that the Llama3 70B model, when used with the search term \texttt{keyphrases}, achieved the highest Precision, $F_1$, $F_1$@5, and $F_1$@10 scores. This suggests that this particular combination is highly effective for concept extraction in this context. On the other hand, the GPT-3.5 Turbo model with the search term \texttt{keywords} showed the highest Recall. This result may be due to the GPT-3.5 Turbo model being trained on datasets similar to Inspec, which could impact the generalizability and fairness of the evaluation when compared to other open-source models. 

Based on the results of Table \ref{tab:zs_evaluations}, the Llama3 70B model was fixed, and term \texttt{keyphrases} is selected. Table \ref{tab:zs_fs_evaluations} evaluates the effect of different zero-shot and few-shot prompt designs. For the \texttt{ZS + Domain} prompt (adding information to only extract keyphrases related to a domain), the Precision, Recall, and $F_1$ scores decreased. This shows that the model focuses on keyphrases specific to the given domain, leading to the exclusion of some relevant keyphrases present in the ground truth but not directly related to the specified domain. For example, when using simple keyphrase extraction, keyphrases such as "graphical user interface" and "scanning data" were extracted. However, when the domain-specific prompt was used, these keyphrases were not extracted. However, this approach ensures that the extracted keyphrases are highly relevant to the specified domain, which can be advantageous depending on the application requirements.

Table \ref{tab:zs_fs_evaluations} illustrates the impact of different context prompts on the evaluation metrics. The incorporation of context into the zero-shot prompts improved the evaluation results. The zero-shot prompt with the task description showed the highest $F_1$, $F_1$@5, and $F_1$@10 scores. This suggests that providing the model with a clear task description enhances its ability to extract relevant keyphrases more accurately.

Additionally, Table \ref{tab:zs_fs_evaluations} presents the evaluation results for different few-shot prompts. The use of random examples improved $F_1$, $F_1$@5, and $F_1$@10 scores, suggesting that variability in the examples enhances the model's generalization and keyphrase extraction performance. The effect of increasing the number of examples was also analyzed. While incorporating one example generally improved performance, adding more examples beyond three did not result in significant improvements. Furthermore, the impact of using the closest documents based on embeddings was evaluated. The few-shot 5-closest prompt achieved the highest Precision score, indicating that carefully selected examples can lead to better precision in keyphrase extraction.
Overall, the differences between the different few-shot prompts are not huge. This shows that it is important to provide an example, but the main advantage is that the model can train on how the output should look and which words are of interest.

Table \ref{tab:combined_evaluation} presents the overall evaluation results, showing the Precision, Recall, $F_1$, $F_1$@5, $F_1$@10, and an average number of extracted keyphrases ($N_{EX}$) for each dataset. The evaluation covers a broad spectrum of models, including EmbedRank \cite{EmbedRank}, MultPAX \cite{MultPAX}, Pyate \cite{PyATE}, RAKE \cite{yake}, FirstPhrases \cite{pke}, KPMiner \cite{KPMiner}, Kea \cite{Kea}, MultipartiteRank \cite{MultipartiteRank}, PositionRank \cite{PositionRank}, SingleRank \cite{SingleRank}, TextRank \cite{TextRank}, TfIdf \cite{tfidf}, TopicRank \cite{TopicRank}, and YAKE \cite{yake}. These models were tested on the Inspec and SemEval2017 datasets. Among them, the few-shot 1-Random prompt using the Llama3 70B model achieved the highest scores in $F_1$, $F_1$@5, and $F_1$@10 metrics. This outcome highlights the crucial role of prompt design and the strategic selection of examples in significantly improving the efficacy of keyphrase extraction. The comprehensive comparison underscores how tailored prompts and example configurations can optimize performance across different datasets and extraction tasks. However, it is noteworthy to mention that our best-performing approach shows lower Recall compared to PositionRank. This suggests that while our method achieves higher precision and top-$k$ performance, it may miss a portion of relevant concepts that graph-based extractors like PositionRank are better at capturing. One possible reason is the conservative behavior of LLMs, especially when using prompts that prioritize correctness, which can lead to under-generation. Future research could explore methods for recall enhancement, such as hybrid models that combine LLM outputs with graph-based methods, or prompt tuning strategies that encourage broader concept retrieval.

\begin{table}
\centering
\caption{Precision, Recall, $F_1$, $F_1$@5, $F_1$@10, and Average Number of Extracted Keyphrases ($N_{EX}$) for each dataset. In the last row, the Few-Shot 1-Random prompt with Llama3 70B model is used as our best ConExion model. Best results are highlighted in bold.}
\label{tab:combined_evaluation}
\resizebox{\textwidth}{!}{
\begin{tabular}{l|rrrrrr|rrrrrr}
\hline
& \multicolumn{6}{c|}{\textbf{Inspec}} & \multicolumn{6}{c}{\textbf{SemEval2017}} \\ \cline{2-13}
\textbf{Model} & \textbf{P} & \textbf{R} & \textbf{\boldmath{$F_1$}} & \textbf{\boldmath{$F_1$@5}} & \textbf{\boldmath{$F_1$@10}} & \textbf{\boldmath{$N_{EX}$}} & \textbf{P} & \textbf{R} & \textbf{\boldmath{$F_1$}} & \textbf{\boldmath{$F_1$@5}} & \textbf{\boldmath{$F_1$@10}} & \textbf{\boldmath{$N_{EX}$}} \\
\hline
MultPax & 0.051 & 0.057 & 0.049 & 0.052 & 0.052 & 5.000 & 0.150 & 0.091 & 0.107 & 0.103 & 0.103 & 5.000 \\
EmbedRank & 0.194 & 0.369 & 0.232 & 0.209 & 0.247 & 8.760 & 0.143 & 0.154 & 0.139 & 0.102 & 0.145 & 9.330 \\
PyateBasics & 0.090 & 0.743 & 0.155 & 0.008 & 0.017 & 47.716 & 0.076 & 0.521 & 0.127 & 0.013 & 0.016 & 62.290 \\
Rake & 0.093 & 0.690 & 0.158 & 0.128 & 0.202 & 46.167 & 0.072 & 0.562 & 0.124 & 0.056 & 0.091 & 72.290 \\
FirstPhrases & 0.164 & 0.586 & 0.242 & 0.193 & 0.227 & 18.457 & 0.139 & 0.331 & 0.186 & 0.086 & 0.146 & 20.0 \\
KPMiner & 0.048 & 0.066 & 0.049 & 0.063 & 0.057 & 6.006 & 0.121 & 0.107 & 0.100 & 0.116 & 0.115 & 7.260 \\
Kea & 0.088 & 0.384 & 0.134 & 0.128 & 0.143 & 20.0 & 0.104 & 0.251 & 0.138 & 0.136 & 0.154 & 20.0 \\
MultipartiteRank & 0.172 & 0.603 & 0.253 & 0.224 & 0.246 & 18.158 & 0.152 & 0.344 & 0.199 & 0.149 & 0.190 & 20.0 \\
PositionRank & 0.166 & \textbf{0.755} & 0.260 & 0.247 & 0.287 & 26.543 & 0.143 & \textbf{0.617} & 0.224 & 0.130 & 0.199 & 39.980 \\
SingleRank & 0.193 & 0.682 & 0.285 & 0.231 & 0.291 & 18.457 & 0.182 & 0.425 & 0.242 & 0.141 & 0.197 & 20.0 \\
TextRank & 0.195 & 0.685 & 0.287 & 0.221 & 0.292 & 18.457 & 0.174 & 0.403 & 0.230 & 0.135 & 0.195 & 20.0 \\
TfIdf & 0.099 & 0.435 & 0.152 & 0.129 & 0.154 & 20.0 & 0.114 & 0.277 & 0.151 & 0.146 & 0.167 & 20.0 \\
TopicRank & 0.169 & 0.531 & 0.242 & 0.230 & 0.241 & 17.074 & 0.145 & 0.340 & 0.192 & 0.152 & 0.187 & 19.870 \\
YAKE & 0.089 & 0.398 & 0.137 & 0.121 & 0.142 & 20.0 & 0.097 & 0.242 & 0.131 & 0.073 & 0.114 & 20.0 \\
\hline
\textbf{FS 1-Random (ours)} & \textbf{0.413} & 0.575 & \textbf{0.451} & \textbf{0.406} & \textbf{0.453} & 7.327 & \textbf{0.301} & 0.372 & \textbf{0.311} & \textbf{0.214} & \textbf{0.302} & 10.840 \\
\hline
\end{tabular}
}
\end{table}

\subsection{Reproducibility}

To ensure reproducibility, 
the versions of datasets and models from HuggingFace were fixed, ensuring that the results could be reproduced. The source code and detailed instructions for running the experiments can be found in our GitHub repository\footnote{\url{https://github.com/ISE-FIZKarlsruhe/concept_extraction}}.

For the baseline models, various models including MultPAX \cite{MultPAX}, PyATE \cite{PyATE}, RAKE \cite{RAKE_library} were employed, alongside several models using the PKE library \cite{pke}, including FirstPhrases, KPMiner, Kea, MultipartiteRank, PositionRank, SingleRank, TextRank, TfIdf, TopicRank, and YAKE. The use of these libraries ensured that all model versions were fixed, facilitating precise replication of our results. 






\section{Conclusion}
\label{sec:Conclusion}

In this paper, an approach for concept extraction from abstract documents using pre-trained large language models (LLMs) was presented. Our method addresses the challenging task of extracting all present concepts related to a specific domain, rather than just the keyphrases summarizing the important information discussed in a document. Comprehensive evaluations of two widely used benchmark datasets demonstrate that our approach performs better than state-of-the-art models. Our emphasis on reproducibility has ensured that the findings have been reliably replicated across various models. One limitation of this work is its reliance on exact lexical matching to filter concepts from the generated output. While this approach ensures that only terms present in the input document are retained, it fails to account for situations where an LLM generates semantically accurate concepts that do not have an exact match in the text. As a result, relevant concepts that capture the meaning are going to be discarded, negatively affecting recall and the overall evaluation.

In the future, we plan to create datasets annotated by domain experts in the field of Materials Science and Engineering (MSE). This will allow us to further test and refine our ConExion models, ensuring they are robust and effective across a broader range of scientific domains.
We also want to explore more methods on how to restrict the LLM models to only produce specific tokens (those that appear in the document). Additionally, we are interested in investigating the effects of continued pre-training and task-specific instruction tuning of LLMs on concept extraction performance. This could provide further insights into how domain adaptation influence model behavior on this task.

\begin{acknowledgments}
  The authors thank the German Federal Ministry of Education and Research (BMBF) for financial support of the project \href{https://www.materialdigital.de/}{Innovation-Platform MaterialDigital} through project funding FKZ no: 13XP5094F (FIZ).
  MatWerk funding acknowledgement
  This publication was written by the NFDI consortium NFDI-MatWerk in the context of the work of the association German National Research Data Infrastructure (NFDI) e.V.. NFDI is financed by the Federal Republic of Germany and the 16 federal states and funded by the Federal Ministry of Education and Research (BMBF) – funding code M532701 / the Deutsche Forschungsgemeinschaft (DFG, German Research Foundation) - project number 460247524.
\end{acknowledgments}

  

\bibliography{sample-ceur}




\end{document}